\documentclass{article}

\usepackage{PRIMEarxiv}

\usepackage[utf8]{inputenc} % allow utf-8 input
\usepackage[T1]{fontenc}    % use 8-bit T1 fonts
\usepackage[hidelinks]{hyperref}       % hyperlinks
\usepackage{url}            % simple URL typesetting
\usepackage{booktabs}       % professional-quality tables
\usepackage{amsfonts}       % blackboard math symbols
\usepackage{nicefrac}       % compact symbols for 1/2, etc.
\usepackage{microtype}      % microtypography
\usepackage{lipsum}
\usepackage{tikz}           %for binary trees
\usepackage{fancyhdr}       % header
\usepackage{graphicx}       % graphics
\usepackage{algorithm}
\usepackage{amsmath}
\usepackage{algpseudocode}
\usepackage[super,numbers,sort&compress]{natbib}
\usepackage{natmove}
\graphicspath{{media/}}  
\graphicspath{{images/}}

\title{Active Learning in Symbolic Regression with Physical Constraints
}

\author{
  \href{https://orcid.org/0000-0003-2174-5557}{\includegraphics[scale=0.06]{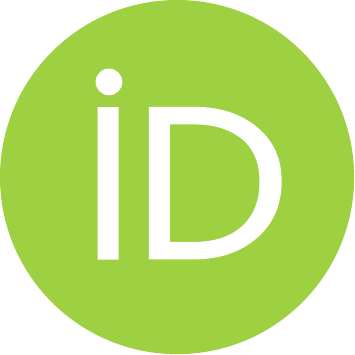}\hspace{1mm}Jorge Medina} \\
	Department of Chemical Engineering \\
	University of Rochester\\
	\texttt{jmedina9@ur.rochester.edu} \\
     \And
\href{}{\includegraphics[scale=0.06]{images/orcid.pdf}\hspace{1mm}Manasi Subhash Gangan}\\
	Department of Biology \\
	University of Rochester\\
	\texttt{mgangan@ur.rochester.edu} \\
       \And
\href{https://orcid.org/0000-000204164-0122}{\includegraphics[scale=0.06]{images/orcid.pdf}\hspace{1mm}Anne S. Meyer}\\
	Department of Biology \\
	University of Rochester\\
	\texttt{anne@annemeyerlab.org} \\
   \And
\href{https://orcid.org/0000-0002-6647-3965}{\includegraphics[scale=0.06]{images/orcid.pdf}\hspace{1mm}Andrew D. White}\thanks{Corresponding author} \\
	Department of Chemical Engineering \\
	University of Rochester\\
	\texttt{andrew.white@rochester.edu} \\  
}

\begin{document}
\maketitle
\begin{abstract}

Evolutionary symbolic regression (SR) fits a symbolic equation to observations, producing a concise, interpretable model. Here we explore using SR to guide which observations to make - namely, guiding choice of experiments based on a set of candidate symbolic equations. We judge this process by its ability to rediscover known equations with as few experiments as possible. 
The proposed observations are chosen with query by committee (QBC), where the committee is a set of plausible symbolic equations. We also consider if adding constraints from prior knowledge to filter equations accelerates rediscovery. The methods perform well, generally requiring fewer observations than deep learning methods. Finally, we use this equation discovery method to learn a multi-dimensional equation describing bacterial growth and discover an accurate Gompertz-like curve.
\end{abstract}

% keywords can be removed
\keywords{Symbolic Regression\and Query by Committee \and Physical Constraints \and Gompertz Model \and Constraints}

\section{Introduction}

A variety of established methods exist for modeling structured data, consisting of a set of features/variables and their corresponding labels. Ranging from traditional machine learning and statistics techniques (linear regression, ridge regression, polynomial regression) to deep learning approaches (neural networks), these methods suffer from constraints and/or interpretability issues, such as limiting the model to a particular shape (e.g., linear), or being too complex to interpret (black box models). Symbolic Regression (SR) is less constrained and searches through the mathematical space of equations to fit the data. SR allows for discovering a broader range of functional relationships, including those with nonlinear or intricate interactions between variables. \cite{SchmidtDistillingfree-form}.

SR is a machine learning technique to propose equations describing data, to fit optimal functional forms and parameters (e.g. equations) \cite{koza_genetic_1994, SchmidtDistillingfree-form}. As the number of features increases, the search space grows exponentially, leading to various strategies to explore it with evolutionary algorithms, sparse models, and neural networks \cite{pysr,ouyang_sisso_2018,udrescu_ai_2020,guimera_bayesian_2020,naik_discovering_2022,schmelzer_discovery_2020}.  SR has been applied successfully to scientific and engineering problems, demonstrating their potential for uncovering real-world mathematical equations.  Weng et al. (2009) utilized SR to propose a model for pipe deterioration, while Baichang et al. (2020) leveraged it to introduce a new descriptor for identifying novel perovskite catalysts. More recently, Li et al. (2023) employed SR to determine electron transfer models of minerals under pressure, among other applications.\cite{MCKAY1997981,SchmidtDistillingfree-form,LACAVA2016292,SANJUAN2020111971,NEUMANN2020123412,li_electron_2023,CAI20064352,reinbold_robust_2021,weng_simple_2020,udrescu_symbolic_2021,berardi_pipes,grundner_data-driven_2023}. 

The complexity of ground-truth equations and noise in the data can complicate the discovery of equations. Fajardo et al. showed modeling data from systems can be split in two: discoverable and non-discoverable ground truth \cite{fajardo-fontiveros_fundamental_2023}. Some ground-truth equations are complex enough that rediscovering them becomes impossible, even in the absence of noise\cite{fajardo-fontiveros_fundamental_2023}. The difficulty in identifying and dealing with noise brings the necessity of devising methods robust to noise to improve SR results. As the ground-truth equation complexity increases, SR methods often demand many more labeled samples, which can incur high costs or require extensive acquisition time. Active learning strategies, such as Query by Committee (QBC), can tackle these challenges by reducing the number of samples needed through informed selection of new experiments to produce labeled data \cite{settles2009active, seung1992}. In this paper, we introduce two straightforward approaches to reduce the data needed for SR. (I) incorporating the QBC strategy into the SR framework, and (II) including soft constraints based on background knowledge in the discovery of equations.

The optimization process may not identify equations with physical meaning; any equation that fits the data could be found. Applying strict restrictions to search only over physically meaningful equations could make finding suitable expressions as challenging as the optimization itself \cite{GoldbergGeneticAlgorithms}. Based on this observation, we decided to use a soft constraint to guide the search in the desired direction, also supported by \cite{GoldbergGeneticAlgorithms}. Various analogous strategies have previously been employed, such as constraining equations to maintain unit correctness \cite{tenachi_deep_2023}, using physical properties to simplify the problem \cite{udrescu_ai_2020,lu_using_2016}, guiding the search for predefined shapes or forms known to be present in the system or deduced from the dataset \cite{CHAKRABORTY2021107470,lu_using_2016}, or checking correctness after equations are rediscovered \cite{fox_incorporating_2023,grundner_data-driven_2023}. Emgle et al, added auxiliary first and second derivatives to be used for constraint optimization \cite{engle_deterministic_2022}. Recently, a soft constrained approach was applied to adsorption equations, revealing that some constraints might not work as expected for genetic algorithms \cite{fox_incorporating_2023}.

We employed the Feynman dataset as a benchmark for comparing the data points required to rederive equations. At its publication, AI Feynman\cite{udrescu_ai_2020} presented state-of-the-art results and is commonly used for performance comparisons. Our findings indicate that our approach frequently outperforms this method. In this work, we focus on a specific application of Genetic Algorithms (GAs) with binary trees as the data structure for representing candidate solutions. The tree representation serves not only as an implementation formalism but also helps in tracking which sub-structures (branches) are favored during evolution. For more details on the workings of genetic algorithms, such as initialization, reproduction, crossover, and mutation, please refer to the Supplementary Information (SI). The rest of this paper explores the efficacy of the active learning technique for rediscovering equations using the incorporation of physical constraints and presents our findings and analysis, finishing with a practical use case on bacterial growth characterization. We confirmed that including physically based constraints improves the generalizability, interpretability, and validity of equations discovered with SR, requiring less data.

% \begin{figure}[h!]
%     \centering
%     \begin{tikzpicture}[level distance=1cm,
%     level 1/.style={sibling distance=3cm},
%     level 2/.style={sibling distance=1.5cm}]
%     \node {+}
%     child {node {*}
%         child {node {A}}
%         child {node {B}}
%             }
%     child {node {-}
%         child {node {$\sin$}
%             child{node{C}}}
%         child {node {D}}
%     };
% \end{tikzpicture}
%     \caption{Binary tree representation of the equation $(A*B)+(\sin(C)-D)$}
%     \label{fig:tree}
% \end{figure}
\section{Theory}
\subsection{Inclusion of Background Knowledge: Physical Constraints}
During optimization, the search is primarily driven by accuracy. However, there is no guarantee that the proposed equation will comply with all the physical constraints of the system, such as divergence, symmetry of variables, or conservation laws. To take advantage of this type of prior knowledge and guide the search towards physically meaningful solutions, the fitness function can be modified to incorporate these constraints as can be seen in Equation \ref{Fitness} where a penalty term is included to balance both accuracy and physical consistency. Given a dataset $(\vec{x},y) \textrm{ of size } N$ and a member of the population $\xi$ the fitness function $f$ is defined as: 

\begin{equation}
    f(\xi,\vec{x},y)= L(\xi,\vec{x},y)+\lambda \phi_k(\xi,\vec{x}) + p_l (\xi)
    \label{Fitness}
\end{equation}

In this equation, $L(\xi,\vec{x},y)$ represents the Root Mean Square Error (RMSE), quantifying the accuracy of each equation over the dataset. The term $\phi_k$ serves as a penalty term based on the constraint $k$, while $\lambda$ is an adjustable parameter set to balance the trade-off between accuracy and physical consistency; for our experiments, we have chosen a value of $\lambda = 100$. A final penalty term is added to target the complexity of the equations, denoted by $p_l$, which promotes size homogeneity in the population. The final fitness or score of the equation $\xi$ is represented by $f(\xi)$. Whenever an equation does not follow physical constraints, it can be penalized with one of the several different approaches described in the SI section \ref{Inclusion_of_background_subsection}.

\subsection{Active Learning} \label{active_learning_subsection}

QBC works by selecting a sample based on the disagreement among a group of experts (the committee). Members of this group should be distinct enough that their disagreement provides valuable information, while being consistently good at describing the system being studied. The algorithm can be described as follows:

\begin{enumerate}
\item Train a model and select a committee.
\item Each member of the committee estimates labels of a non-labeled pool.
\item Measure and select the point with the highest disagreement between members from the pool.
\item Label it and repeat until necessary.
\end{enumerate}

\begin{figure}[h!]
    \centering
    \includegraphics[width=10cm]{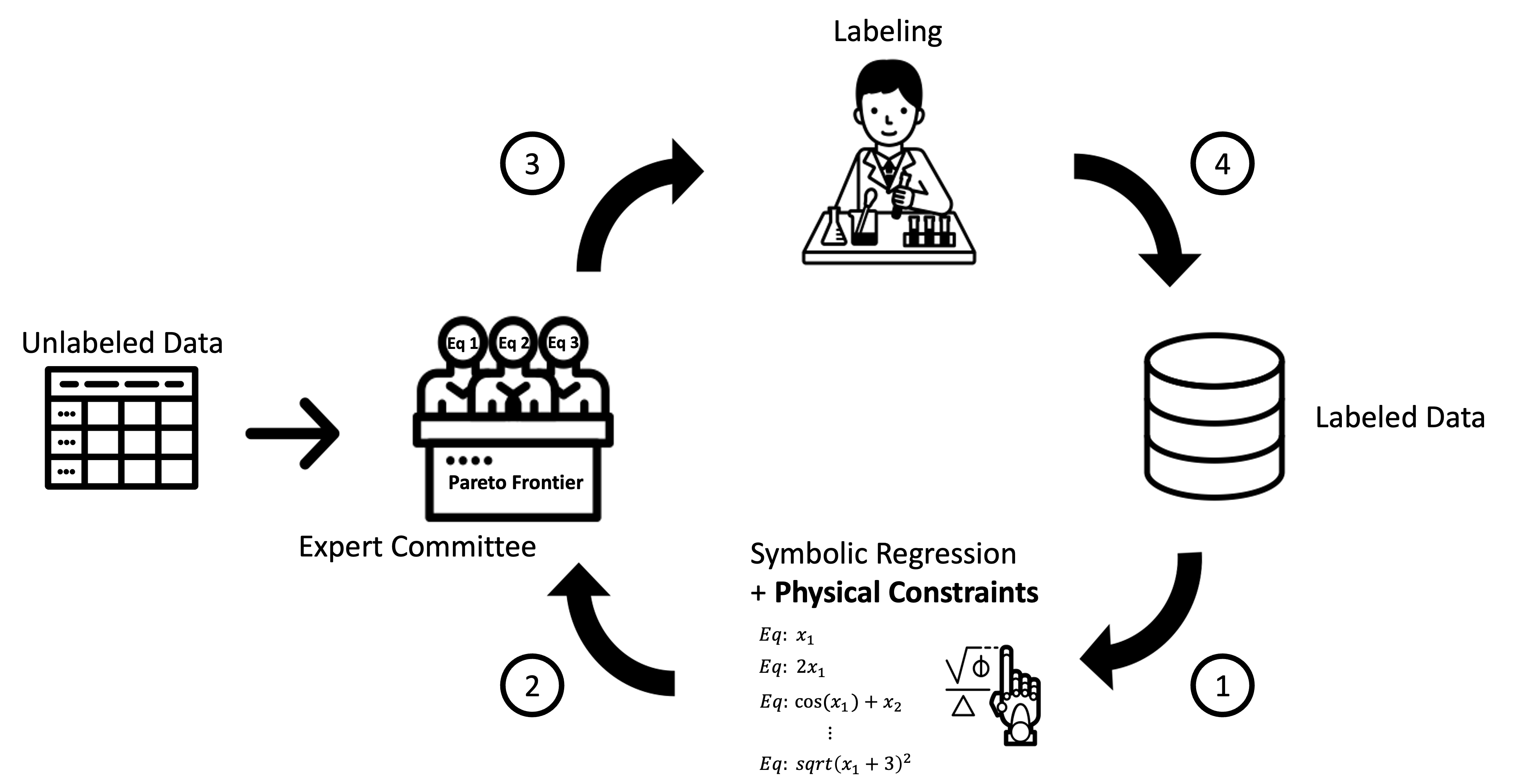}
    \caption{Query By Committee Depiction. 1) Symbolic Regression with physical constraints 2) outputs a Pareto frontier of equations that act as expert models, and measures disagreement from unlabeled data to 3) select the most informative point to 4) label and iterate.}
    \label{fig:QBCDepiction}
\end{figure}

Disagreement is defined in terms of an unlabeled dataset $D_u$, and $N$ committee members $K = \{\Xi_{1},\Xi_{2},....,\Xi_{N}\}$ Each member $i$ makes a prediction on every unlabeled point $j$, denoted as $\Xi_{i}(\vec{x_j})=Y_{ij}$. For this study, committee members are selected from the Pareto frontier. These members represent the trade-off between the accuracy and complexity of the equations and generally provide valuable insights into the whole domain of the system. A depiction of the process can is shown in Figure \ref{fig:QBCDepiction}

 Comparable studies in the literature yielded the best results when applied to low complexity equations, where active learning setups were deemed unnecessary since rediscovery occurred with minimal data, less than ten. However, the approach faced limitations when dealing with higher complexity equations.\cite{haut_active_2022}. Bongard et al. discussed the implementation of QBC to generate and sample informative training examples in a grammatical inference problem \cite{JMLR:v6:bongard05a}. Murari et al. used confidence intervals of models obtained through SR to guide experimentation towards out-of-distribution tests that could falsify the fitted models \cite{Murari_SR_AL_with_Confidence_intervals}.

\paragraph{Measurement of Disagreement} Measuring disagreement $d(K,\vec{x})$ involves the unsampled dataset ($\vec{x}$) and the committee members ($K$). Two measures of disagreements were tested. The first one is the coefficient of variation (CV), which is the ratio of the standard deviation ($\sigma$) to the mean ($\mu$) of the predicted labels by all members of the committee. This offers a comprehensible metric for assessing the dispersion of the predictions:

\begin{equation}
\mu(K,\vec{x}) = \frac{1}{N} \sum_{i=1}^{N} \Xi_{i}(\vec{x})
\end{equation}

\begin{equation}
\sigma(K,\vec{x}) = \sqrt{\frac{1}{N} \sum_{i=1}^{N} (\Xi_{i}(\vec{x}) - \mu(K,\vec{x}))^2}    
\end{equation}

\begin{equation}
 d_{CV}(K,\vec{x}) = \frac{\sigma(K,\vec{x})}{\mu(K,\vec{x})}    
\end{equation}

The coefficient of variation facilitates the comparison of dispersion across different data points or scales.

The second measure is an information-based measure of disagreement (IBMD) \cite{IBMD}, which takes into account the relative differences between pairs of predictions:

\begin{equation}
    d_{IBMD}(K,\vec{x}) = \frac{1}{|C^N_2|}\sum\limits_{(i,j)\in C^N_2} \log\left(\frac{|\Xi_{i}(\vec{x})-\Xi_{j}(\vec{x})|}{\max(\Xi_{i}(\vec{x}),\Xi_{j}(\vec{x}))}+1\right) 
    \label{IBDM}
\end{equation}

where $(\xi_i,\xi_j)$ represents a pair of equations from the committee, and $|C^N_2|$ is the number of possible pair combinations of $N$ members. The IBMD measure calculates the logarithm of the ratio between the absolute difference of the predictions and the maximum of the two predictions, incremented by 1. These logarithmic values are then averaged for all possible pair combinations according to Equation \ref{IBDM}.

\section{Methodology}

SymbolicRegression.jl or PySR \cite{pysr} is a Julia package designed for discovering mathematical equations through regularized evolution. This high-performance package offers an efficient implementation of SR, enabling the identification of underlying mathematical relationships within complex data sets. 
Equation representations are based on binary trees, with numerous hyperparameters for customization. The same parameters were chosen for all experiments to ensure fair and straightforward comparisons across multiple cases. These primarily consisted of default parameters for the evolutionary process. The maximum number of iterations was adjusted to 100, and the selected binary operators included addition ($+$), subtraction ($-$), multiplication ($*$), and division ($/$). The chosen unary operators were inverse (inv), square, cube, exponent ($\exp$), square root ($\sqrt()$), and cosine ($\cos$). Restrictions were applied to prevent nested operations, such as $\cos(\cos(x))$. The software's simplicity, efficiency, and ability to customize loss functions made it a suitable choice for our research. 

Initially, to assess our methodology, we leveraged the observation that rediscovering an equation often leads to a marked enhancement in accuracy (more than 10 orders of magnitude for noiseless systems), accompanied by a minimal rise in equation complexity. This facilitates an automated process for configuring experiments and asserting success using loss metrics. Nevertheless, noise makes less evident the rediscovery of the ground truth, especially when the noise source or level is unknown. It is essential to employ a more hands-on approach for success evaluation, which involves visually examining whether the equation has been successfully rediscovered.

In the initial stage, our primary goal was to demonstrate the effectiveness of our method through a comprehensive toy case. To examine the impact of incorporating physical constraints, we explored the one-dimensional gravity equation:$\frac{Gm_1m_2}{r^2}$. All variables were held constant (number of iterations, operator set, default evolution parameters), except the fitness function. We acknowledge that, given sufficient time (e.g., 100 iterations), the algorithm will inevitably rediscover the equation for this equation. Therefore, we limited the number of iterations to 20, making the task more challenging for the algorithm.

Secondly, to evaluate the performance of Query By Committee, we set up an active learning scenario involving the following equation: $f = (1 / 2\pi)^{1/2} \exp(-(x_1 - x_2) / \sigma)^2 / 2$ from the Feynman dataset. We started with three data points and examined the percentage of rediscovery as new data points were added based on random sampling, QBC with IBMD, or CV. This step-wise approach allowed us to compare both approaches systematically. 

Next, we benchmarked the model against AIFeynman, an SR framework that used the Feynman equations dataset \cite{udrescu_ai_2020}. This comparison enables us to assess the performance of our approach relative to existing methods, highlighting its strengths and potential weaknesses.

Then, we continued with stress tests, introducing two distinct levels of noise into the data. This stage aims to assess the robustness of our findings when faced with real-world scenarios where data has noise from experiments.

Finally, we use the methodology in a practical use case, bacterial growth characterization. And use the results to parameterized known growth curves models to the data. \textit{Shewanella oneidensis} MR-1, an electrogenic marine microorganism that can use various chemical species including oxygen as a terminal electron acceptor for respiration \cite{Shewanella}. Although various variables such as temperature, pH, oxygen availability, mutations, types of electron donor, and carbon source type and its concentration could influence the study, we simplified the study by focusing solely on the effects of lactate concentration. This setting involves dealing with different types of noise ( epistemic and aleatoric) from the measurement equipment and the complex nature of living systems. We varied the lactate concentration from 0 to 100 mM, starting at a pH of 7 and keeping constant room temperature throughout the experiments. Initially, we tested 11 equally spaced lactate concentrations. Subsequently, QBC was employed to guide through to the next experiments.

After initial experimentation with the methodology, it was decided to only use the physical constraints for a limited number of steps during the evolutionary process. After this initial period, the constraints were removed, allowing exploration of a wider range of potential solutions. This approach aimed to strike a balance between guiding the model towards physically meaningful solutions and giving the algorithm freedom to take full advantage of the evolutionary operations with a head start in the right direction. Figure \ref{fig:subtrees expl} shows the effectiveness of this approach.

\section{Results}
As the research methodology was structured into three phases: (I) proof of concept, (II) benchmarking against AI Feynman, and (III) conducting robustness analysis, it enabled a systematic evaluation of our proposed SR approach before the method was tested against real data.

\subsection{Physical Constraints}
To evaluate the effectiveness of incorporating physical constraints and Query By Committee in SR, a proof of concept study was conducted using the well-known gravitational force equation $F = \frac{Gm_1m_2}{r^2}$ as a test case. This equation was chosen due to its simplicity and multiple possible constraints, providing a clear benchmark for assessing the impact of the proposed methods.

Four types of constraints were considered for this experiment: divergence, asymptotic, monotonicity, and positivity. The goal was to investigate how each of these constraints affect the ability of SR  to rediscover the gravitational force equation. A pool of 100 points for $m_1$, $m_2$, and $r$ was used, and randomly sampling 10 of them for each experiment. 

\begin{figure}[h!]
    \centering
    \includegraphics[width=9cm]{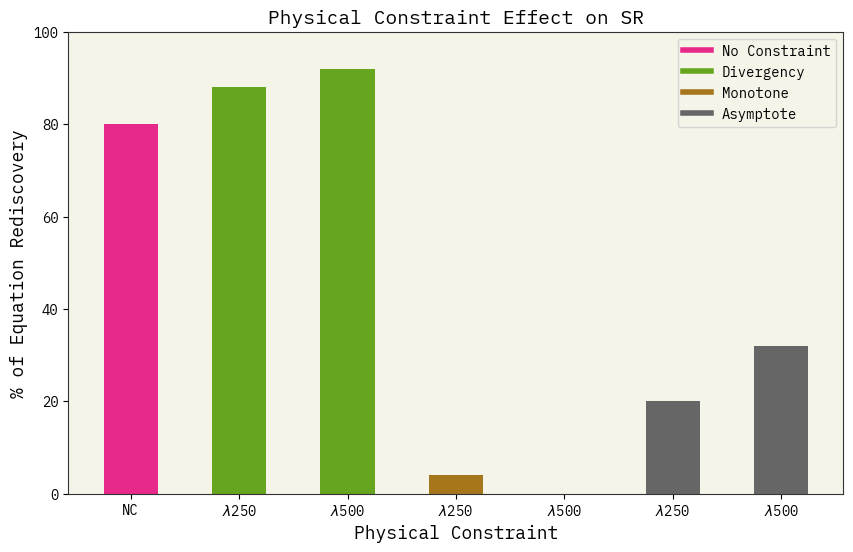}
    \caption{Rediscovery of Gravitational Law with different types of constraints. $\lambda$ regulates the physical constraint strength during optimization. NC: No Constraint}
    \label{fig:Gravity}
\end{figure}

Out of the four constraints examined, SR incorporating divergence constraints in the loss function demonstrated superior performance compared to non-constrained regression. This improvement led to an approximately 10\% increase in the rediscovery rate, as illustrated in Figure \ref{fig:Gravity}. Conversely, the remaining three constraints resulted in reduced performance. Interestingly, certain constraints were observed to hinder the search for the optimal direction, suggesting that even soft constraints may degrade the optimization process. Divergence can be obtained with the term $1/r$, while other constraints may require more intricate structures, such as larger sub-trees or alternative available options that may differ from the ground truth but still technically comply with the constraints.

\subsection{Query By Committee}

\begin{figure}[h]
    \centering
   % \begin{subfigure}{\textwidth}
        \centering a)
        \includegraphics [width=0.41\linewidth, height=0.2\textheight]{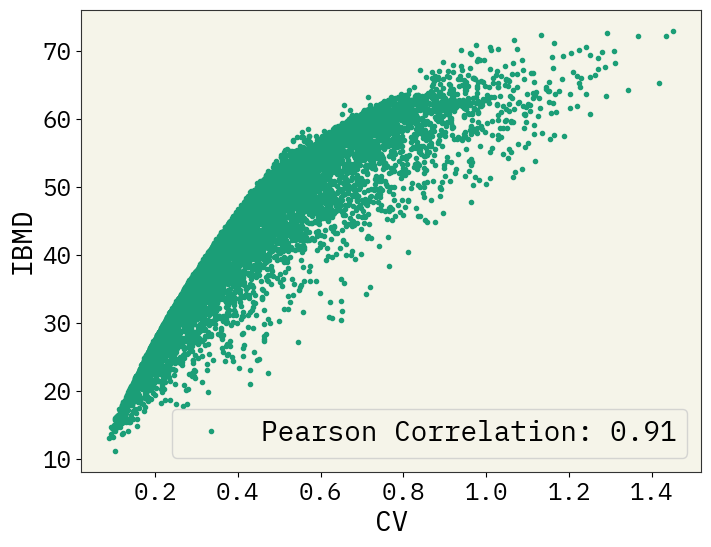}%
        \hfill b)
        \includegraphics [width=0.41\linewidth, height=0.2\textheight]{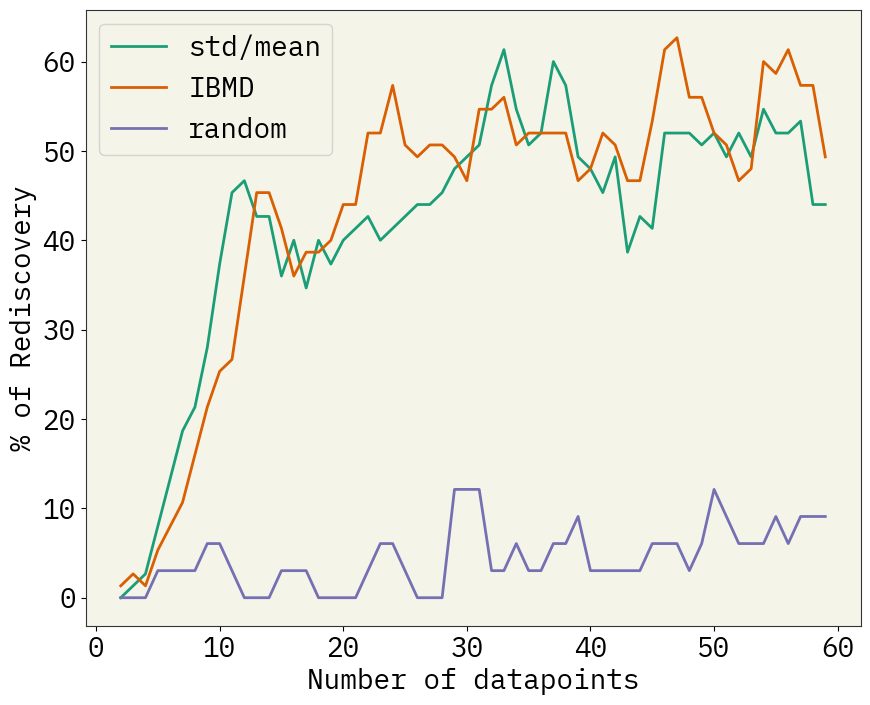}
   % \end{subfigure}
        \caption{Evaluation of disagreement measures. a) Disagreement analysis of 15 equations from PySR and 10,000 data points, resulting in a correlation coefficient of 0.91. Similar points exhibit maximum disagreement. b) Comparison of $f = (1 / 2\pi)^{1/2} \exp(-(x_1 - x_2) / \sigma)^2 / 2$  rediscovery performance versus the number of data points when adding new labeled points through active learning and random addition}
    \label{fig:disagreements}
\end{figure}

\begin{figure}[h!]
    \centering
    \includegraphics[width=10cm]{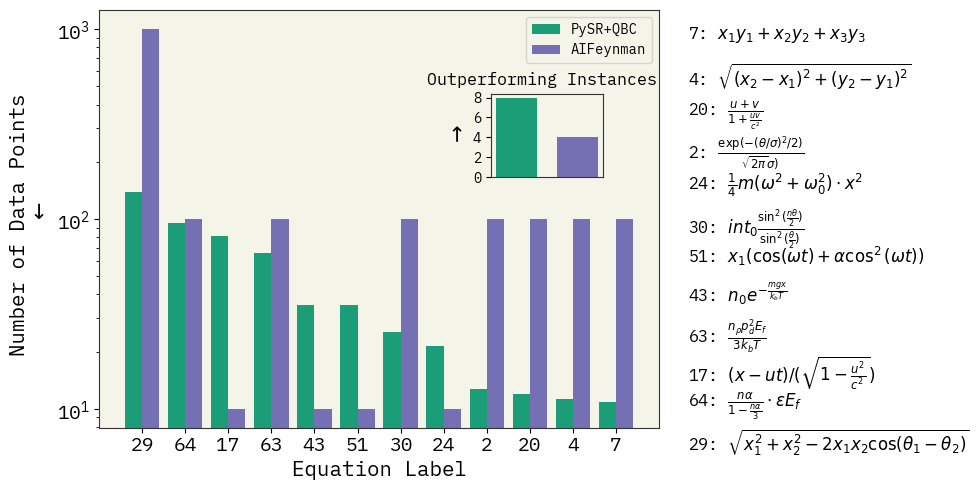}
    \caption{Number of samples needed for rediscovery of non-trivial equations of Feynman dataset with (green) and without (blue) constraints. The inset chart shows that from the twelve tested equations, PySR with QBC outperformed AIFeynman in eight of them. Arrows show the direction of improvement.}
    \label{fig:PYSRVSFEYNMAN}
\end{figure}

We compare both measures of disagreement, mentioned in section \ref{active_learning_subsection}, and they appear to work similarly. As shown in Figure \ref{fig:disagreements}a, comparing the disagreement between the equations on the Pareto frontier using either method (CV or IBMD) yielded comparable outcomes. A correlation coefficient close to one was anticipated, since both measurements represent the variability between the predictions from each equation. Figure \ref{fig:disagreements}b demonstrates that employing QBC leads to the more frequent rediscovery of the same equation using fewer data points compared to merely adding new data at random. Consequently, the remaining experiments presented in Figure \ref{fig:PYSRVSFEYNMAN} were conducted using QBC with IBMD.

The findings presented in Figure \ref{fig:PYSRVSFEYNMAN} demonstrate that our model, with the implementation of QBC, outperforms AIFeynman in 8 out of 12 test cases in rediscovering equations using fewer data points. However, it is important to note that AIFeynman's results were only reported in orders of magnitude, which limited our ability to make a direct comparison between the precise number of data points required for each model.

\subsection{Robustness Experiments}

Three distinct equations, selected for their non-triviality and inherent physical constraint, were utilized to assess the method both in noiseless and noisy setups. The noise level was increased to the point where no rediscovery was obtained. In these cases, convergence was rare and not measured enough to obtain valuable statistics, results are either successful (Rediscovery) or unsuccessful (No Rediscovery). Table \ref{rediscoverywithnoise} shows another comparison between un-constrained and constrained optimization where using constraints can rediscover the ground truth in a `high' noise level.

\begin{figure}[h!]
    \centering
    \includegraphics[width=9cm]{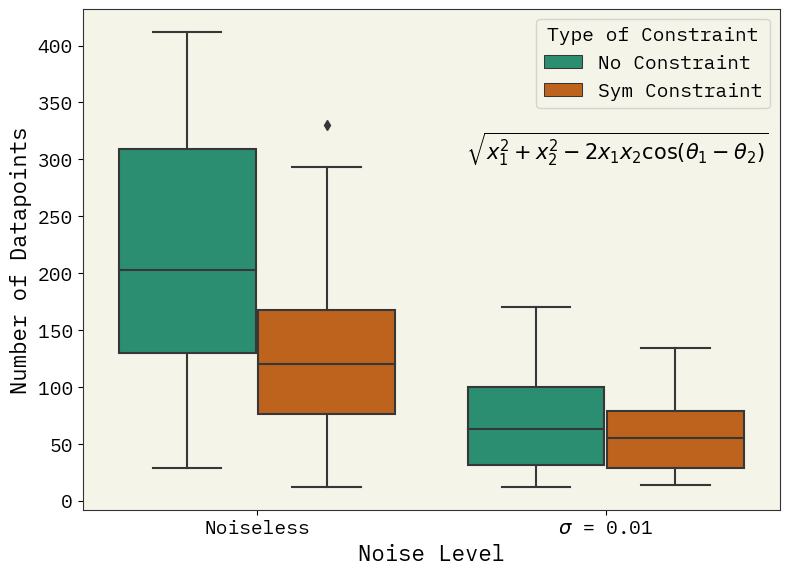}
    \caption{Illustrating the Effects of Noise on Rediscovery: A Comparison between Constrained and Unconstrained Optimizations. The figure presents results for the expression $\sqrt{x_{1}^{2} + x_{2}^{2} - 2x_{1}x_{2}\cos(\theta_{1} - \theta_{2})}$. P-values are 0.061 for noiseless conditions and 0.443 for a noise level of 0.01. The joint p-value for comparing unconstrained and symmetry-constrained optimization is 0.127. }
    \label{fig:boxplot1}
\end{figure}

The first equation, as depicted in Table \ref{results1} and Figure \ref{fig:boxplot1}, showed an enhancement in rediscovery by employing symmetry constraints (p-value of 0.061). Interestingly, and against our expectations, the performance exhibits improvement as noise is added. However, this can be attributed to the inherent limitations of conducting a fully automated process with noise. Increasing the level of noise, some iterations often converged prematurely (without rediscovery of the ground truth equation), resulting in a higher number of runs with fewer data points. Nevertheless, an improvement can be observed in the box plot for both noise experiments.

Similar analysis on two different equations (see SI) shows less improvement with constrained optimization. Comparing the complexity of three equations reveals that for more complex equations, constraints have positive effects on optimization. 

\begin{table}[h]
\centering
\caption{Rediscovery of $\sqrt{x_{1}^{2} + x_{2}^{2} - 2x_{1}x_{2}\cos(\theta_{1} - \theta_{2})}$}
\begin{tabular}{lcccc}
\toprule
Type of Experiment   & Noiseless  & 0.01 Noise & 0.1 Noise \\
\midrule
No Constraint  &  213      &     71.20    &  No Rediscovery          \\
Symmetry       &  139      &   59.19      & No Rediscovery         \\

\bottomrule
\label{results1}
\end{tabular}
\end{table}

\subsection{\textit{Shewanella oneidensis} Growth Characterization}
Bacterial growth, or many other types of growth, is usually modeled and characterized with parametric sigmoid/logistic functions with respect to time \cite{Gompertz_models, Richards_Model}. Growth analysis usually involves a set of fixed variables, measured over time. SR can allow more flexibility and the exploration of other variables without constraining the final model, which can allow the discovery of interpretative relations between said variables and the model's parameters.

\textit{Shewanella}, a bacterium known for its extracellular respiration. Understanding the environmental factors that alter its growth can elucidate ways to optimize systems that use this bacteria for hydrogen production or bio-electric systems \cite{PNAS_SHEWANELLA,Shewanella_bioelec,Shewanella}. Even though multiple variables can be controlled and observed, we initially focused on the carbon source concentration.

Bacterial growth typically exhibits four distinct phases: lag, exponential, stationary, and death. Our study focuses on the first three phases, specifically examining how the lag time, growth rate, and carrying capacity of bacterial growth vary with the concentration of the carbon source lactate \cite{Buchanan_Whiting_Damert_1997}. Our approach allows studying how different features impact the growth parameters simultaneously, instead of most standard approaches where each variable must be analyzed separately or have to assume a mathematical growth model to fit the data \cite{Wang_Fan_Chen_Terentjev_2015,Wang_Chen_Terentjev_2015}.

To start, we analyzed bacterial growth using a series of concentrations ranging from 0 to 100 mM lactate. The optimization incorporated an asymptotic physical constraint, defined as in definition \ref{asymptote_cons}, where n and c are hyperparameters, that module how strict the constraint is during optimization. For this experiment $c=0.5$, and $n=500$. Table \ref{table_parameters_bact} shows other parameters used during optimization.

After gathering initial data, QBC indicated that the next region worth exploring is [0-20], this is further detailed in Supplementary Figure \ref{fig:BactFirstIteration}. Within the first range, we observed distinct growth behaviors: rapid increase from 0-20 mM, suggestive of diauxic\footnote{Diauxic growth, meaning double growth, indicates that after the the growth settles a second carbon source starts a second growth phase.} shifts between 30-60 mM, and a pseudo-stationary phase from 60-100 mM, where increases in lactate concentration do not alter growth dynamics. 

\begin{figure}[h!]
    \centering
    \includegraphics[width=17.5cm]{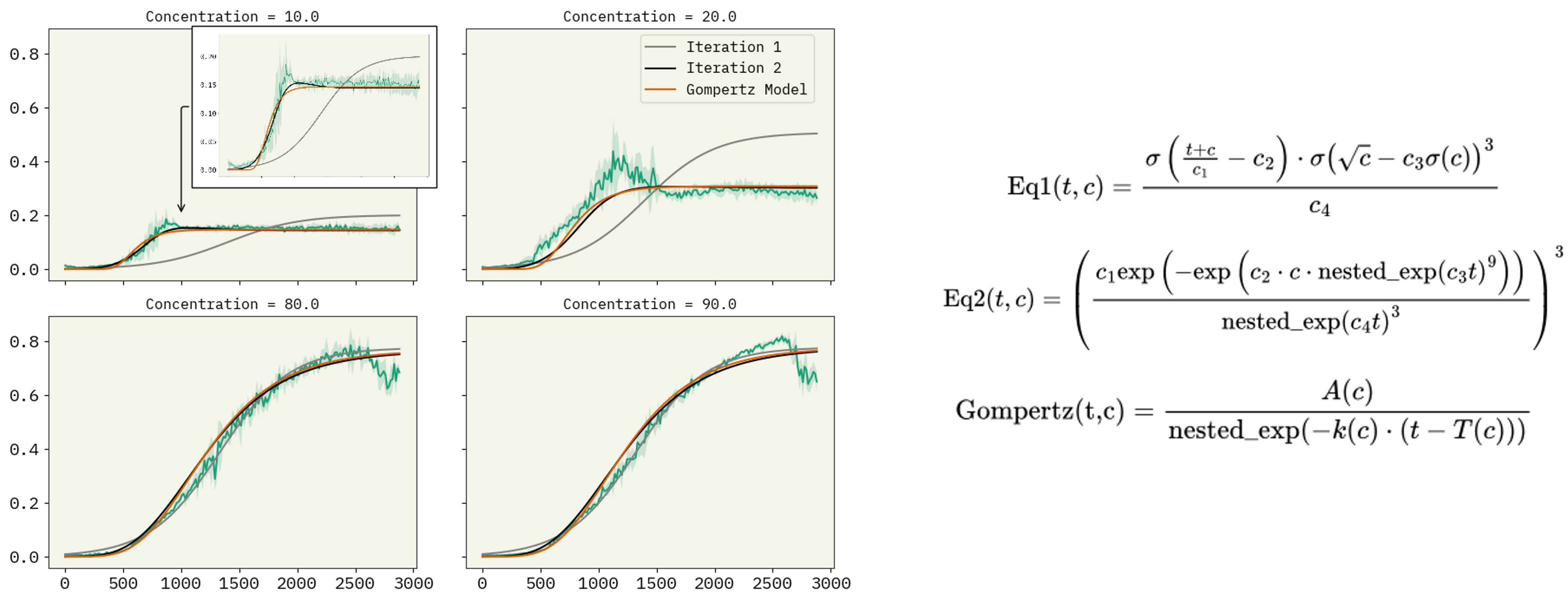}
    % \caption{Right: Improvement in accuracy with addition of data selected by the Pareto frontier disagreement in four different lactate concentrations. Left: Equations with highest accuracy at each iteration, and the interpretable equation derived from the best fit. $\sigma = \frac{1}{1+\exp(-x)}$, $\text{nested_exp}(x)=\exp\exp(x)$, $A(c)$, $T(c)$ and $k(c)$ are derived in the SI }
        \caption{
    Right: Improvement in accuracy with addition of data selected by the Pareto frontier disagreement in four different lactate concentrations. 
    Left: Equations with highest accuracy at each iteration, and the interpretable equation derived from the best fit. $c_i$ represent constants,
    $\protect\sigma = \frac{1}{1+\exp(-x)}$, 
    nested\_exp$(x)=\exp(\exp(x))$, 
    $A(c)$, 
    $T(c)$ and 
    $k(c)$ are derived in the SI 
    }
    \label{fig:panelsandequation}
\end{figure}

In the second iteration, we conduct growth tests at ten equally spaced lactate concentrations ranging from 0 to 20 mM, plus an additional concentration at 1 mM to complete the set of eleven experiments feasible per batch. This denser sampling within the 0-20 mM range, not only resulted in more accurate equations, it also allowed for a more detailed interpretation of growth responses with changes in the concentration of Lactate at low ranges (see Supplementary Information Equations \ref{overfittedeq}-\ref{growth_ratetotal}). If we decided to continue to the next iteration the Pareto frontier disagrees the most around 15 mM. It then becomes an experimental choice, if extra precision can be achieved, experimenting near the 15mM range can be done, if not, the next most informative region goes from [70,90] mM of lactate. In figure \ref{fig:panelsandequation} four different concentrations are displayed together with the best equations at each iteration, this equations don't follow any known growth model, but a Gompertz model can be derived (see SI).

% \begin{figure}[h]
%     \centering
%    % \begin{subfigure}{\textwidth}
%          \centering a)
%         \includegraphics[width=0.45\linewidth, height=0.31\textheight]{ScoresIt1v2.png}
%         \label{fig:bactqbc1}
%         \hfill b)
%          \includegraphics[width=0.45\linewidth, height=0.31\textheight]{ScoresIt2v2.png}
%          \label{fig:bactqbc2}
%    % \end{subfigure}
%         \caption{Disagreement scores after a) first and b) second experimental iterations. After the first round, the concentration region with the most disagreement is  [0,20]mM, while on the second iteration the \textit{new} region suggested to explore is around 15 nM
%         }
%     \label{fig:scoresdisagreements}
% \end{figure}

\section{Conclusion and Discussion}
This study presents a simple approach to use Symbolic Regression for active learning by integrating QBC and physical constraint regularization. The proposed method prioritizes informative samples using the Pareto frontier of candidate equations and guides the algorithm toward physically meaningful expressions by incorporating physical constraints as regularization terms.

Our results demonstrate that the application of QBC facilitates the rediscovery of equations using fewer data points than AI Feynman. Moreover, incorporating symmetry as a constraint in symbolic regression showed improvement against a non-constrained regression. The removal of constraints needs to be explored more in-depth to find the most appropriate strength of constraints without hindering the evolutionary process. 

The proposed method exhibits resilience against noise, successfully rediscovering equations under various noise levels for specific cases. By integrating QBC and physical constraint regularization, this approach provides a framework for uncovering meaningful mathematical expressions across a wide range of applications, even with limited data. 

On the topic of noisy data, the method is useful for the real case scenario, with a considerable level of noise, for \textit{Shewanella} growth curves characterization. It allowed us to describe how the growth rate, capacity, and lag time parameters relate to changes in concentration with interpretable equations.

\section*{Code and Data Availability}
The SymbolicRegresion.jl package was forked from its github repo \cite{pysr} and its changes can be found at: \url{https://github.com/Jgmedina95/SRwPhysConsWL}. The methodology employed here can be tested at \url{https://aldemo.fly.dev/demo}

\section*{Conflicts of Interest}
Authors have no conflicts of interest to declare

\section*{Acknowledgments}
We thank the Center for Integrated Research Computing (CIRC) at the University of Rochester for providing computational resources and technical support.

 This work  has been supported by funds from the Robert L. and Mary L. Sproull Fellowship gift and U.S. Department of Energy, Grant No. DE-SC0023354.

%Bibliography
\bibliographystyle{unsrt}  
\bibliography{references}  

\newpage
\appendix

\section{Supp. Material}
\renewcommand\thefigure{\thesection.\arabic{figure}}
\renewcommand\thetable{\thesection.\arabic{table}}
\setcounter{figure}{0}  
\setcounter{table}{0}

\subsection{Additional Theory}

Genetic Algorithms (GAs) are computational methods inspired by the principles of natural selection, incorporating processes such as reproduction, mutation, and crossover. They involve a population of candidate solutions, referred to as individuals, which consist of components known as genes. Individuals can be represented using various data structures, such as strings, trees, or stacks. Common terms used in GA literature include parents, children, genes and fitness.\cite{GoldbergGeneticAlgorithms}

The fitness assigns a score to each individual based on how well they perform in solving the target problem. This score is then used to guide the selection process, allowing fitter individuals to have a higher probability of contributing their genetic material to the next generation, thereby promoting the evolution of increasingly effective solutions over time. \cite{GoldbergGeneticAlgorithms}
\subsubsection{Initialization, Reproduction, and Crossover in Genetic Algorithms}

The more members in the population, the broader the search and the better chance of success, but at the expense of computation time. A random combination of unary and binary operators (e.g., $\sin$, $\cos$, inv(x), $+$, $-$, $*$), variables, and constants are formed to get an initial population. After this, ``evolution'' can start.

Once the first generation is initialized, reproduction takes place. Reproduction is the process where the `fittest' members of the population get passed towards the next generation. To decide which member is better, a metric called fitness, considering both accuracy and complexity, is used. The fitness metric can be the Mean Absolute Error or Root Mean Square Error, and complexity is the number of tree nodes. Although it can be customized as needed. 

After reproduction, the crossover comes next. Crossover, also known as recombination, is a primary genetic operator in genetic algorithms that generates new offspring solutions by combining the features (genes) of two parents \cite{holland_adaptation_1992,GoldbergGeneticAlgorithms}. Inspired by the natural process of sexual reproduction, crossover helps promote genetic diversity and exploration of the search space, which is crucial for solving complex optimization problems \cite{DeJong}. This allows searching outside the original population on the mathematical equation space. This is represented in figure \ref{fig:crossover_depiction}\\
\begin{figure}[h]
    \centering
\begin{tikzpicture}[level distance=1cm,
level 1/.style={sibling distance=2.0cm},
level 2/.style={sibling distance=1.5cm}]

\node at (-1, 0) {Before};
\node at (7, 0) {After};
% First tree
\node at (0,0) {+}
  child {node {*}
    child {node {A}}
    child {node {B}}
  }
  child {node {-}
    child {node {C}}
    child {node {D}}
  };

% Second tree
\node at (4,0) {*}
  child {node {+}
    child {node {E}}
    child {node {F}}
  }
  child {node {-}
    child {node {G}}
    child {node {H}}
  };

% New tree after crossover (from first tree)
\node at (8,0) {+}
  child {node {*}
    child {node {A}}
    child {node {B}}
  }
  child {node {-}
    child {node {G}}
    child {node {H}}
  };

% New tree after crossover (from second tree)
\node at (12,0) {*}
  child {node {+}
    child {node {E}}
    child {node {F}}
  }
  child {node {-}
    child {node {C}}
    child {node {D}}
  };

\draw[thick] (6,0.5) -- (6,-2.5);

% Draw boxes around the child nodes
\draw[blue, thick] (2.00,-0.82) rectangle (0.00,-2.25);
\draw[red, thick] (5.90,-0.82) rectangle (4,-2.25);
\draw[red, thick] (8,-0.82) rectangle (10,-2.25);
\draw[blue, thick] (12,-0.82) rectangle (14,-2.25);
\end{tikzpicture}

    \caption{Crossover depiction between two members of the population}
\label{fig:crossover_depiction}
\end{figure}

Despite being predominantly heuristic-based, genetic algorithms benefit from the ``schema theorem" \cite{GoldbergGeneticAlgorithms}, which formalizes the underlying intuition and provides a theoretical foundation for their effectiveness in optimization and search tasks. Introduced by David E. Goldberg \cite{GoldbergGeneticAlgorithms}, the schema theorem asserts that genetic algorithms inherently operate on multiple schemata concurrently. These schemata, which serve as building blocks or templates, consist of shared characteristics or patterns found in various solutions within the population. According to the schema theorem, the evolutionary process intrinsically rewards advantageous traits (higher fitness) while penalizing undesirable traits (lower fitness) among these schemata. Through iterative selection, crossover, and mutation, genetic algorithms progressively converge towards optimal or near-optimal solutions. Consequently, the schema theorem elucidates the principles behind genetic algorithms, demonstrating their ability to efficiently navigate the search space and identify high-quality solutions to intricate optimization challenges.

\subsubsection{Mutation: a safeguard}
Mutation, see Figure \ref{fig:mutation_depiction} is another fundamental genetic operator in genetic algorithms that helps to maintain diversity within the population and prevent premature convergence to suboptimal solutions \cite{GoldbergGeneticAlgorithms,Mitchell1996AnIT} It is inspired by the natural process of mutation in biological organisms, where random changes occur in an individual's genetic material. In the context of genetic algorithms, mutation introduces small random alterations to the candidate solutions, or chromosomes, which can lead to the exploration of new points in the search space and thus facilitate the discovery of better solutions.

The mutation step is typically applied after the crossover operation, and its probability is often much lower than that of crossover \cite{DeJong} There are various mutation operators, and their choice depends on the problem's representation and domain. For example, in a binary-encoded genetic algorithm, a commonly used mutation operator is bit-flip mutation, which involves flipping the value of a randomly selected bit (changing 0 to 1 or vice versa) in a chromosome \cite{holland_adaptation_1992}. 

\begin{figure}[h]
    \centering
\begin{tikzpicture}[level distance=1cm,
level 1/.style={sibling distance=2.0cm},
level 2/.style={sibling distance=1.5cm}]

% Before and After labels
\node at (-1, 0) {Before};
\node at (3, 0) {After};

% Tree before mutation
\node at (0,0) {+}
  child {node {*}
    child {node {A}}
    child {node {B}}
  }
  child {node {-}
    child {node {C}}
    child {node {D}}
  };

% Tree after mutation
\node at (4,0) {+}
  child {node {*}
    child {node {A}}
    child {node {B}}
  }
  child {node {-}
    child {node {C}}
    child {node {X}} % Changed node
  };
\draw[thick] (2,0.5) -- (2,-2.5);
% Draw boxes around the nodes before and after mutation
\draw[red, thick] (1.95,-1.80) rectangle (1.55,-2.20); % Before mutation
\draw[blue, thick] (5.95,-1.80) rectangle (5.55,-2.20); % After mutation

\end{tikzpicture}
    \caption{Mutation depiction of a member of the population}
\label{fig:mutation_depiction}
\end{figure}

\subsection{Inclusion of background knowledge}
\label{Inclusion_of_background_subsection}
In the following pseudocode, we present an algorithm that incorporates a penalty term into the loss evaluation process to account for constraint violations. This concise representation demonstrates how to combine the objective function and penalty term using a weight parameter, lambda. By including the penalty term, the algorithm can optimize solutions while adhering to background knowledge and constraints.

\begin{algorithm}
\caption{Inclusion of a Penalty in Loss Evaluation}
\begin{algorithmic}[1]
\Procedure{EvaluateWithPenalty}{population, objectiveFunc, penaltyFunc, $\lambda$}
    \For{each individual in population}
        \State loss $\gets$ objectiveFunc(individual)
        \State penalty $\gets$ penaltyFunc(individual)
        \State individual.fitness $\gets$ loss + $\lambda$ penalty
    \EndFor
\EndProcedure
\label{algorithm}
\end{algorithmic}
\end{algorithm}

Additionally, here we describe some forms of physical constraints: 

Divergence.\ Is the behavior of a function as its input approaches a particular point. In some cases, it may be necessary to ensure that the function does not diverge (i.e., does not become infinite). For example, in chemical engineering applications, it might be important to guarantee that the concentration of a reactant remains finite. A simple way to represent a divergence constraint is as follows:

\begin{equation}
\phi_{\textrm{div}}(\xi,a) = \left\{
    \begin{array}{ll}
        0, &  |\xi(a)| = \infty, \\
        1, &  |\xi(a)| \neq \infty.
    \end{array}
\right.
\end{equation}

Symmetry.\ In this context, refers to the property of a function where permuting two variables does not change its predictions (e.g., $\xi(x_1, x_2) = \xi(x_2, x_1)$) or any example where commutativity applies. Symmetry can be particularly relevant in cases where the order of variables should not affect the system's behavior.

\begin{equation}
\phi_\textrm{symm}(\xi,\vec{x}) = \left\{
\begin{array}{lr}
0, & \xi(x_1,x_2) = \xi(x_2,x_1) \\
1, & \xi(x_1,x_2) \neq \xi(x_2,x_1)
\end{array}
\right.
\end{equation}

Sign Constraints (Positivity/Negativity). \
In some systems, measured properties are always positive (or negative) within a certain domain. For example, this could be relevant when dealing with concentrations, which should always be non-negative. A simple way to represent sign constraints is as follows:

\begin{equation}
\phi_\textrm{sign}(\xi,\vec{x}) = \left\{
\begin{array}{lr}
0, & \xi(\vec{x}) \geq 0  \quad  \\
1, & \textrm{otherwise}
\end{array}
\right.
\end{equation}

Monotonicity Constraints (Increasing/Decreasing Functions).
In some systems, the relationship between variables can be strictly increasing or decreasing within a certain domain. For example, this could be relevant when studying the relationship between temperature and pressure in an ideal gas under constant volume, where temperature and pressure are directly proportional. Monotonicity constraints can be represented based on the sign of the derivative as follows:

\begin{equation}
\phi_\textrm{mono}(\xi,\vec{x}_i) = \left\{
\begin{array}{lr}
0, & \frac{d\xi(\vec{x})}{dx_i} \geq 0 \quad  \\
1, & \textrm{otherwise}
\end{array}
\right.
\end{equation}

This constraint enforces that the function is either strictly increasing or, flipping the inequality, strictly decreasing (non-positive derivative) within the specified domain or feature.

Asymptotic Constraints. Asymptotic behavior can be described by the first derivative as it approaches zero. Growth studies and catalytic systems contain asymptotic limits. The following equation represents an asymptotic constraint, in which $c>0$ and $n>0$ are hyper parameters that modulate the strength of the constraint. 

\begin{equation}
\phi_\textrm{asymptote}(\xi,\vec{x}_i) = \left\{
\begin{array}{lr}
0, & \left|\frac{d\xi(\vec{x})}{dx_i}\right| \leq c \quad  \\
n\left(\left|\left|\frac{d\xi(\vec{x})}{dx_i}\right| - c\right|\right), & \textrm{otherwise}
\end{array}
\right.
\label{asymptote_cons}
\end{equation}

In addition to these constraints, chemical engineering applications may require the incorporation of other types of physical constraints, such as conservation laws, boundary conditions, or monotonicity constraints, depending on the specific problem being addressed.

\section{Additional Results}
\begin{figure} [h]
    \centering
    \includegraphics[width=16cm,height=5cm]{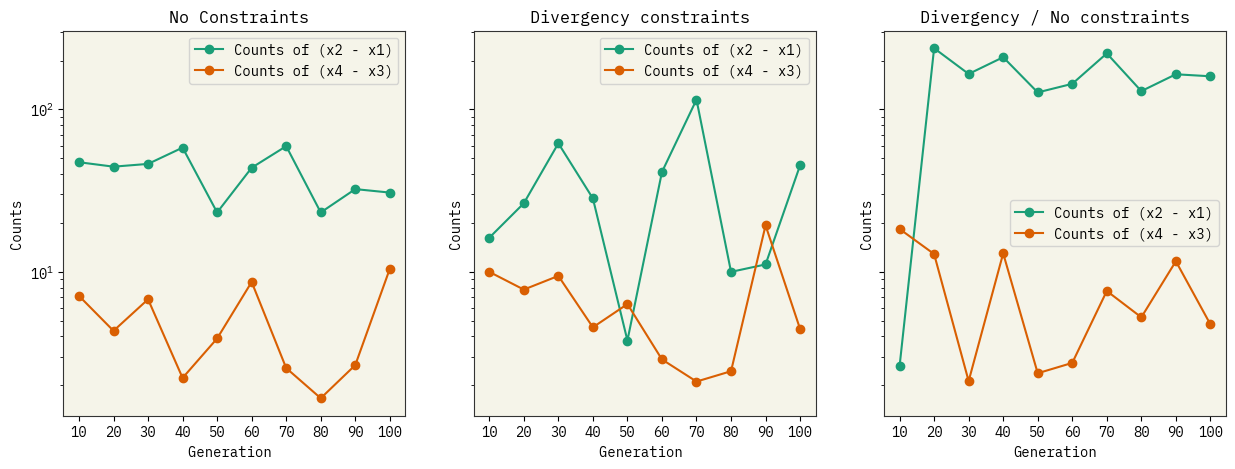}
    \caption{Counts of substructures $(x_{2} - x_{1})$ and $(x_{4} - x_{3})$ for equation  $\frac{1}{\\sqrt{(x_{2} - x_{1})^{2} + (x_{4} - x_{3})^{2}}}$. Results from explorations without rediscovery. It is an indirect assessment of the optimization process, the higher the count, the more indication that the search is performed in the right part of the search space. Without constraints, first panel, substructure counts are very homogeneous but in small amounts. With constraints, substructures seem to appear higher but in a more chaotic way, and a mix of constrained and unconstrained evolution seems to outperformed both methods.  }
    \label{fig:subtrees expl}
\end{figure}

We provide evidence of our approach with constraints with the subtree counts on the rediscovery of equation $\frac{1}{\\sqrt{(x_{2} - x_{1})^{2} + (x_{4} - x_{3})^{2}}}$. As can be seen in the Figure \ref{fig:subtrees expl}

\begin{figure}[h!]
    \centering
    \includegraphics[width=9cm]{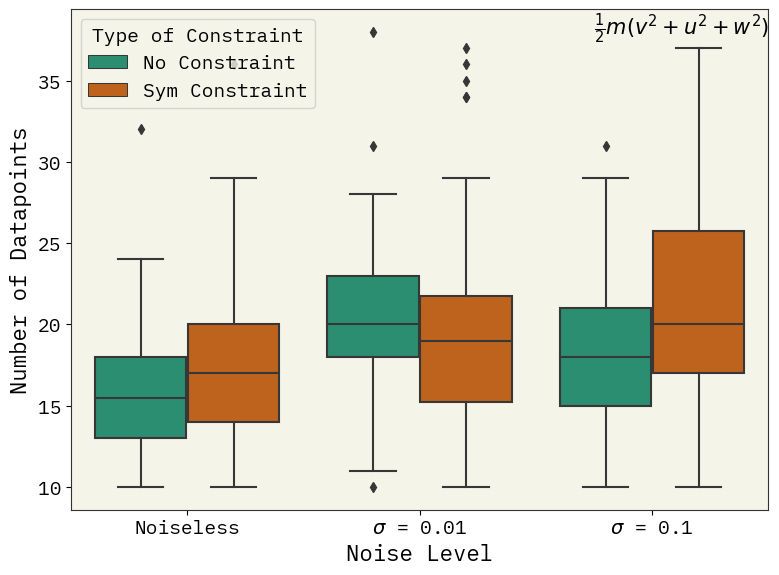}
   \caption{Illustrating the Effects of Noise on Rediscovery: A Comparison between Constrained and Unconstrained Scenarios. The figure presents results for the expression $\frac{1}{2}m(v^{2} + u^{2} + w^{2})$. P-values are 0.6350 for noiseless conditions, 0.0007 for a noise level of 0.01, and 0.005 for a noise level of 0.1. The joint p-value for comparing unconstrained and constrained optimization under the influence of noise is 0.0002.}
    \label{fig:kinenergyresults}
\end{figure}

We provide tables and plots that illustrate the impact of noise on the rediscovery of equations and the effectiveness of implementing physical constraints. Tables \ref{tablekinenergy} and \ref{rediscoverywithnoise} present the results for the rediscovery of $\frac{1}{2}m(v^{2} + u^{2} + w^{2})$ and $\frac{1}{\\sqrt{(x_{2} - x_{1})^{2} + (y_{2} - y_{1})^{2}}}$, respectively, under different noise levels and constraint types. Figures \ref{fig:kinenergyresults} and \ref{fig:boxplot3} visually represent these results, providing a clearer comparison between the use of physical constraints and no constraints in the equation rediscovery process.

\begin{table}[h]
\centering
\caption{Rediscovery of $\frac{1}{2}m(v^{2} + u^{2} + w^{2})$}
\begin{tabular}{lcccc}
\toprule
Type of Experiment &Noiseless  & 0.01 Noise & 0.1 Noise \\
\midrule
No Constraint    &15.61      & 20.1        &  18.33          \\
Symmetry         &17.4       & 19.6     & 21.16         \\

\bottomrule
\end{tabular}
\label{tablekinenergy}
\end{table}

The results from the third equation, illustrated in Table \ref{rediscoverywithnoise} and Figure \ref{fig:boxplot3}, do not exhibit any enhancement when compared to unconstrained optimization. In fact, the p-values indicate that a significant difference exists between the constrained and unconstrained methods when no constraints are applied favoring the latter, showing the opposite behaviour. However, when the noise level is increased to 0.1, the ground truth is only rediscovered in the constrained scenario.

\begin{table}[h]
\centering
\caption{Rediscovery of $\frac{1}{\\sqrt{(x_{2} - x_{1})^{2} + (y_{2} - y_{1})^{2}}}$}
\begin{tabular}{lcccc}
\toprule
Type of Experiment  & Noiseless & 0.01 Noise & 0.1 Noise \\
\midrule
No Constraint & 11           & 18        &  No Rediscovery           \\
Symmetry     & 16          & 23      & Rediscovery         \\
Divergency   & 16         & 23       & Rediscovery         \\
\bottomrule
\label{rediscoverywithnoise}
\end{tabular}
\end{table}

\begin{figure}[htb!]
    \centering
    \includegraphics[width=9cm]{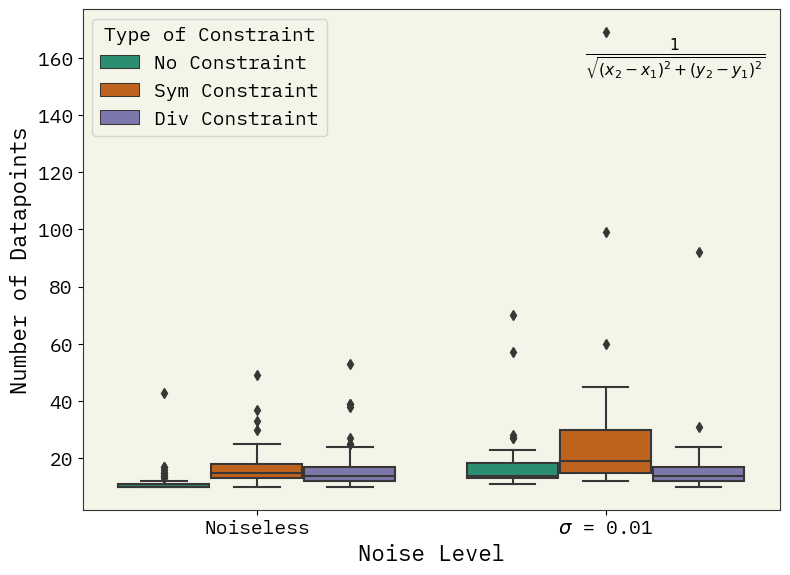}
  \caption{Illustrating the Effects of Noise on Rediscovery: A Comparison between Constrained and Unconstrained Scenarios. The figure presents results for the expression $\frac{1}{\\sqrt{(x_{2} - x_{1})^{2} + (y_{2} - y_{1})^{2}}}$. P-values for Symmetry vs No Constraint are 0.0266 for a noise level of 0.01 and 1.22e-11 for noiseless conditions; for Div vs No Constraint, p-values are 0.3319 for a noise level of 0.01 and 3.98e-10 for noiseless conditions. The joint p-value for comparing constrained and unconstrained optimization under the influence of noise is 1.04e-18.}
    \label{fig:boxplot3}
\end{figure}
\clearpage

\subsection{\textit{Shewanella onodeinesis} Growth}
\label{Shewanella_odeinesis}
Bacterial growth is well known and can be fitted by a family of expressions (like sigmoids and logistic)\cite{Richards_Model,Gompertz_models}. Different expressions found in literature are usually reparametrized versions of themselves, but they usually have in common three to five parameters that define the growth curve. The three parameters that we are interested in this study are Capacity (or maximum asymptote), the lag time, or the time the system requires to adapt to the new environment and start growing, and the growth rate, which defines how fast the system grows.

The experiments for \textit{Shewanella} were performed in triplicate with the following conditions and equipment 
\begin{table}[h]
\centering
\caption{Bacterial Growth Characterization Experimental Parameters \\
$^a$ Every experiment is tripled, from which the statistics ($\mu$  and $\sigma$) are obtained. From a Gaussian($\mu,\sigma$), ten points are sampled at each time.}
\begin{tabular}{lcccc}
\toprule
\textbf{Parameter} & \textbf{Settings} \\
\midrule
Unary Functions & \{$\exp()$ ,  $  \ln$ ,$ \sqrt()$ ,$    ()^2$, $()^3$ , $\frac{1}{1+\exp(-x)}$ , $\exp(-\exp(x))$\} \\
Binary Functions &  \{$  +$ , $  -$ , $  x$ , $  \div$\} \\
Number of Populations &  $100$ \\
Type of Constraint & Asymptote at $t>3500 \text{min}$ \\
Iterations with Constraint & $10$ \\
Iterations without Constraint & $90$\\
Total Number of Iterations & $100$ \\
Data Augmentation\textsuperscript{a} &  10 point augmentation\\
Equipment &    BioTek Synergy H1 Microplate Reader           \\
Temperature & 30°C\\
Growth Media &      Minimal media \cite{PNAS_SHEWANELLA} supplemented with lactic acid under aerobic conditions              \\
\bottomrule
\end{tabular}

\label{table_parameters_bact}
\end{table}

As an example of using Symbolic Regression to give meaningful, interpretable expressions, starting from the second Pareto-frontier, see Table \ref{second-pareto-frontier}, we are going to reach a more amenable expression. The equation with the lowest error is:
\begin{equation}
y\left(t,c\right)=\frac{\left(\frac{0.925026}{N\left(-0.0566564c\cdot N\left(-0.002074753t\right)^{9}\right)}\right)^{3}}{N\left(-0.002074753t\right)^{9}}
\label{overfittedeq}
\end{equation}
where 
\begin{equation}
    N(x) = \exp(\exp(x))
\end{equation}
From equation \ref{overfittedeq}, we can sample asymptote and lag time at different concentrations and then fit this equation to a more understandable Gompertz model, see Equation \ref{Gompertz model}, where each parameter is modeled from equations \ref{lag_time}$-$\ref{growth_ratetotal}:
\begin{equation}
    y(t,c) = A{c}\exp(-\exp(-k_g{c}(t-T_l{c})))
    \label{Gompertz model}
\end{equation}

\begin{equation}
    T_l\left(c\right)=420+(1065-420)\exp(-\exp(-0.05806(c-18.97684)))
    \label{lag_time}
\end{equation}
\begin{equation}
    A\left(c\right)\ =\ 0.78937\exp\left(-\exp\left(-0.065\left(c-16.807\right)\right)\right)
\end{equation}

 $k_g$ could be fitted by two different expressions fitting different growth rates behaviors at different concentration ranges. The use of sigmoid $s(x) = \frac{1}{1+exp(-x)}$ is used to model the ``step'' behaviour between $k_1$ and $k_2$ around $c=35 \text{mM}$
\begin{equation}
    k_{1}\left(c\right)=-0.00601\exp\left(-\exp\left(0.35\left(-0.4c+6.75\right)\right)\right)
\end{equation}
\begin{equation}
    k_{2}\left(c\right)=-0.00601\exp\left(-\exp\left(0.3\left(-1.18c+15.75\right)\right)\right)+0.00831
\end{equation}
\begin{equation}
    k_g\left(c\right)\ =\ \left(1-s\left(c-35\right)\right)k_{1}\left(c\right)+s\left(c-35\right)k_{2}\left(c\right)
    \label{growth_ratetotal}
\end{equation}

The resulting equations demonstrate how symbolic regression can effectively and interpretably fit a new variable (lactate concentration) into a known physical model. Here, it successfully models three growth parameters independently of time, unlike traditional methods that fit each curve to a separate Gompertz model.

\begin{figure}[h]
    \centering
    \includegraphics[width=15cm]{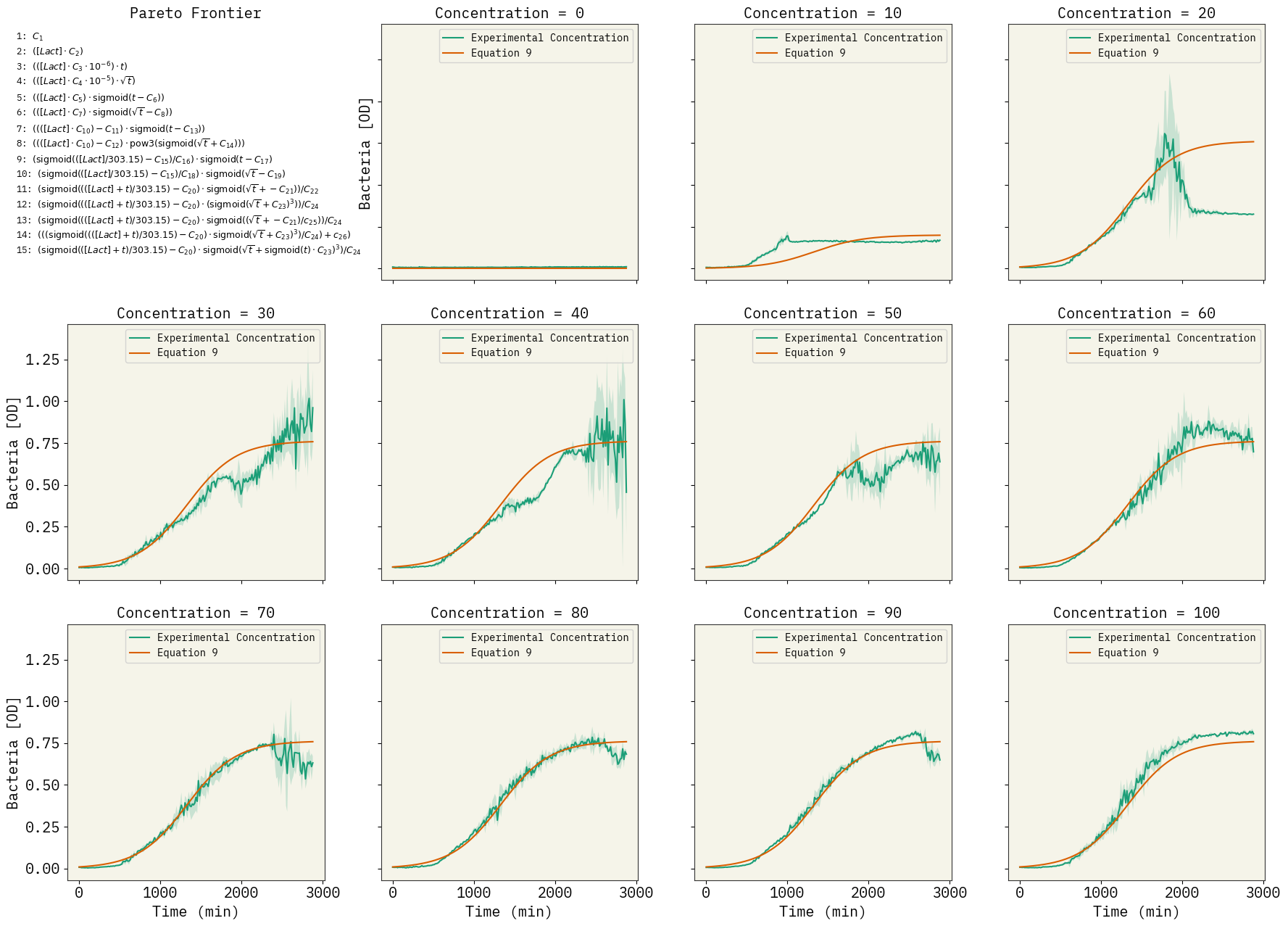}
  \caption{First Iteration of Experiments for Bacterial Growth with different lactate concentrations.
  Top Left. Pareto Frontier of equations of the system. The experimental data is plotted together with one of the equations. It can be seen in the plots that the equation curve changes with concentration up to 20 mM and then remains constant.}
    \label{fig:BactFirstIteration}
\end{figure}

\begin{figure}[h!]
    \centering
    \includegraphics[width=15cm]{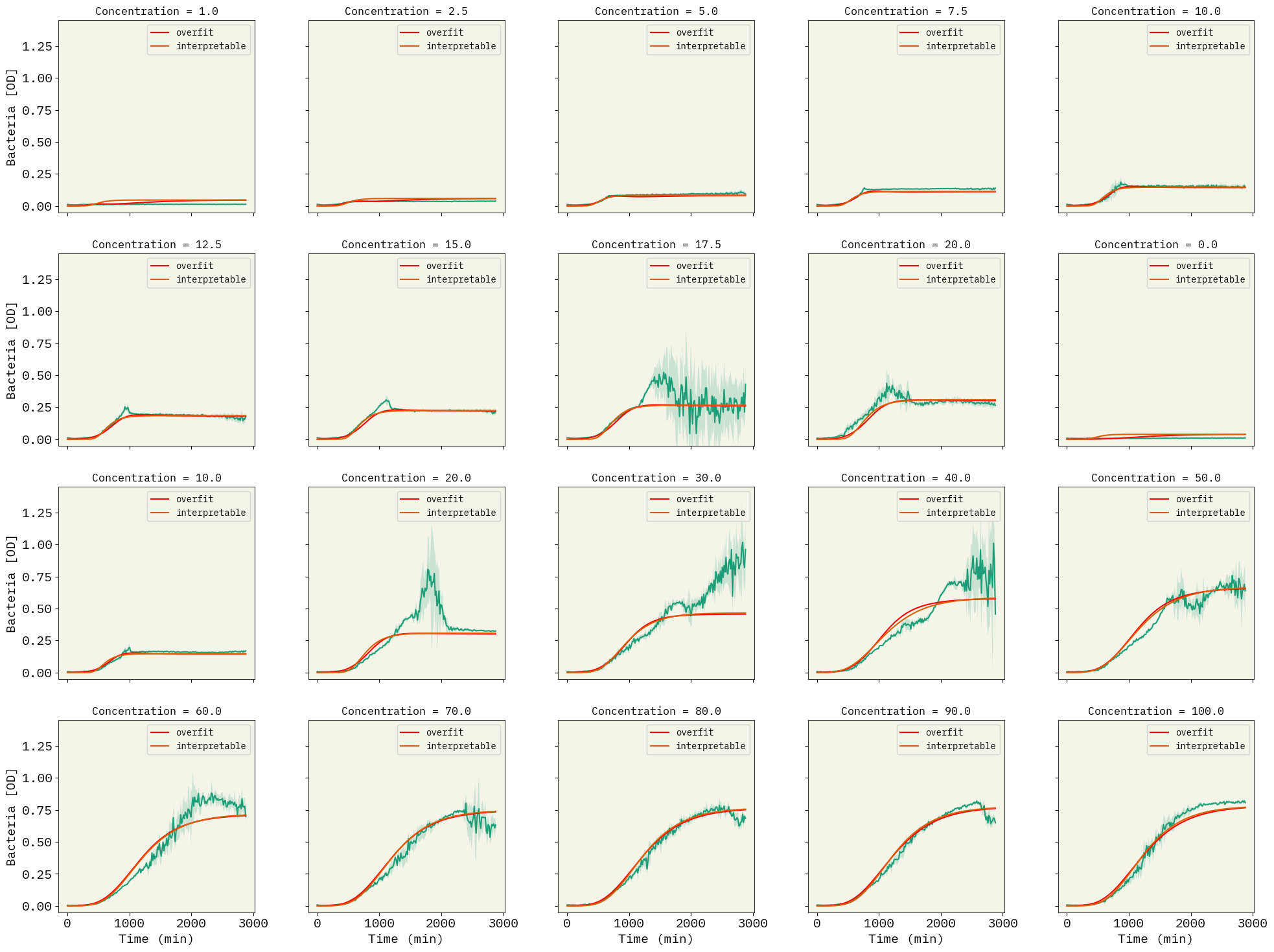}
  \caption{Second Iteration of Experiments for Bacterial Growth with different lactate concentrations.
  The experimental data is plotted together with two of the equations. One is equation \ref{overfittedeq} and equation \ref{Gompertz model} from above. It can be appreciated how the low and high concentration regimes are fitted by the same equation.}
    \label{fig:boxplot3}
\end{figure}

\clearpage
\begin{table}[h!p]
\caption{Pareto Frontier of the second optimization}
\begin{tabular}{cccl}
\hline
Index & Complexity & Loss & Equation \\
\hline
1 & 1 & 0.061588 & $$0.212884$$ \\
2 & 3 & 0.040335 & $$(c * 0.005238)$$ \\
3 & 4 & 0.036347 & $(\sqrt c * 0.045676)$ \\
4 & 5 & 0.010394 & $(c * 3.788\text{e-}6) * t$ \\
5 & 6 & 0.007008 & $(\sqrt c * 3.264\text{e-}5) * t$ \\
6 & 8 & 0.006426 & $((\sqrt c - 0.7062) * t) * 3.627\text{e-}5$ \\
7 & 9 & 0.006359 & $(\sqrt{(c - \sqrt c)} * t) * 3.615\text{e-}5$ \\
8 & 10 & 0.006106 & $((\sqrt{ c - 0.7782)} * (t - c)) * 3.819\text{e-}5$ \\
10 & 11 & 0.006040 & $(\sqrt{(c - \sqrt c)} * (t - c)) * 3.628\text{e-}5$\\
11 & 12 & 0.005471 & $c * \left(((t * 4.235e-6) + 0.009731) - 0.001173\sqrt c\right)$ \\
12 & 13 & 0.003449 & $\left(0.9250 \exp(-\exp(-0.07138c)) / \mathrm{nested\_exp}(-0.002074t)^3\right)^3$ \\
13 & 15 & 0.003292 & $\left(0.9336 \exp(-\exp(-0.07038c)) / \mathrm{nested\_exp} ( -0.001830(t - c))^2\right)^3$ \\
14 & 16 & 0.002752 & $\left(0.9314 \exp(-\exp(-0.06353c))*\exp\left(-\exp( -0.002031(t - \frac{c}{0.1438}))\right)^3\right)^3 $\\
15 & 18 & 0.002357 & $\left(\frac{0.9314 \exp(-\exp(-0.06353c))} { \mathrm{nested\_exp}(-0.002031t * \mathrm{nested\_exp}(-0.06353c))^3}\right)^3$ \\
16 & 19 & 0.002239 & $\left(\frac{0.9250 \exp(-\exp(-0.05666c))}{\left( -0.002075\mathrm{nested\_exp}(t * \mathrm{nested\_exp}((-0.02601c)^3))\right)^3}\right)^3$ \\
17 & 20 & 0.002115 & $ \left(\frac{0.9250 \exp(-\exp(-0.05666c*\mathrm{nested\_exp}(-0.002075t)^9))}{ \mathrm{nested\_exp}(-0.002075t)^3}\right)^3$  \\
\hline
\end{tabular}
\label{second-pareto-frontier}
\end{table}
\setcounter{figure}{0}  
\setcounter{table}{0}

\end{document}